\documentclass[10pt,twocolumn,letterpaper]{article}

\usepackage{iccv}
\usepackage{times}
\usepackage{epsfig}
\usepackage{graphicx}
\usepackage{amsmath}
\usepackage{amssymb}

\usepackage[accsupp]{axessibility} 

\usepackage[pagebackref=true,breaklinks=true,letterpaper=true,colorlinks,bookmarks=false]{hyperref}

\iccvfinalcopy 

\usepackage{booktabs}
\usepackage{xcolor}
\usepackage{makecell}
\usepackage{pifont}

\definecolor{darkgreen}{RGB}{0,127,0}
\definecolor{darkred}{RGB}{200,0,0}
\def\greencheckmark{\textcolor{darkgreen}{\checkmark}}
\def\redxmark{\textcolor{darkred}{\ding{55}}}

\usepackage[capitalize]{cleveref}
\crefname{section}{Sec.}{Secs.}
\Crefname{section}{Section}{Sections}
\Crefname{table}{Table}{Tables}
\crefname{table}{Tab.}{Tabs.}

\newcommand{\mytabular}[1]{\centering\scalebox{0.75}{#1}}

\def\iccvPaperID{1809} 

\begin{document}
%
%
%
%
%
%
%
\newcommand{\ba}{{\mathbf{a}}}
\newcommand{\bb}{{\mathbf{b}}}
\newcommand{\bc}{{\mathbf{c}}}
\newcommand{\bd}{{\mathbf{d}}}
\newcommand{\bolde}{{\mathbf{e}}}
\newcommand{\boldf}{{\mathbf{f}}}
\newcommand{\bg}{{\mathbf{g}}}
\newcommand{\bh}{{\mathbf{h}}}
\newcommand{\bi}{{\mathbf{i}}}
\newcommand{\bj}{{\mathbf{j}}}
\newcommand{\bk}{{\mathbf{k}}}
\newcommand{\bl}{{\mathbf{l}}}
\newcommand{\bm}{{\mathbf{m}}}
\newcommand{\bn}{{\mathbf{n}}}
\newcommand{\bo}{{\mathbf{o}}}
\newcommand{\bp}{{\mathbf{p}}}
\newcommand{\bq}{{\mathbf{q}}}
\newcommand{\br}{{\mathbf{r}}}
\newcommand{\bs}{{\mathbf{s}}}
\newcommand{\bt}{{\mathbf{t}}}
\newcommand{\bu}{{\mathbf{u}}}
\newcommand{\bv}{{\mathbf{v}}}
\newcommand{\bw}{{\mathbf{w}}}
\newcommand{\bx}{{\mathbf{x}}}
\newcommand{\by}{{\mathbf{y}}}
\newcommand{\bz}{{\mathbf{z}}}

\newcommand{\bA}{\mathbf{A}}
\newcommand{\bB}{\mathbf{B}}
\newcommand{\bC}{\mathbf{C}}
\newcommand{\bD}{\mathbf{D}}
\newcommand{\bE}{\mathbf{E}}
\newcommand{\bF}{\mathbf{F}}
\newcommand{\bG}{\mathbf{G}}
\newcommand{\bH}{\mathbf{H}}
\newcommand{\bI}{\mathbf{I}}
\newcommand{\bJ}{\mathbf{J}}
\newcommand{\bK}{\mathbf{K}}
\newcommand{\bL}{\mathbf{L}}
\newcommand{\bM}{\mathbf{M}}
\newcommand{\bN}{\mathbf{N}}
\newcommand{\bO}{\mathbf{O}}
\newcommand{\bP}{\mathbf{P}}
\newcommand{\bQ}{\mathbf{Q}}
\newcommand{\bR}{\mathbf{R}}
\newcommand{\bS}{\mathbf{S}}
\newcommand{\bT}{\mathbf{T}}
\newcommand{\bU}{\mathbf{U}}
\newcommand{\bV}{\mathbf{V}}
\newcommand{\bW}{\mathbf{W}}
\newcommand{\bX}{\mathbf{X}}
\newcommand{\bY}{\mathbf{Y}}
\newcommand{\bZ}{\mathbf{Z}}

\newcommand{\calA}{{\mathcal{A}}}
\newcommand{\calB}{{\mathcal{B}}}
\newcommand{\calC}{{\mathcal{C}}}
\newcommand{\calD}{{\mathcal{D}}}
\newcommand{\calE}{{\mathcal{E}}}
\newcommand{\calF}{{\mathcal{F}}}
\newcommand{\calG}{{\mathcal{G}}}
\newcommand{\calH}{{\mathcal{H}}}
\newcommand{\calI}{{\mathcal{I}}}
\newcommand{\calJ}{{\mathcal{J}}}
\newcommand{\calK}{{\mathcal{K}}}
\newcommand{\calL}{{\mathcal{L}}}
\newcommand{\calM}{{\mathcal{M}}}
\newcommand{\calN}{{\mathcal{N}}}
\newcommand{\calO}{{\mathcal{O}}}
\newcommand{\calP}{{\mathcal{P}}}
\newcommand{\calQ}{{\mathcal{Q}}}
\newcommand{\calR}{{\mathcal{R}}}
\newcommand{\calS}{{\mathcal{S}}}
\newcommand{\calT}{{\mathcal{T}}}
\newcommand{\calU}{{\mathcal{U}}}
\newcommand{\calV}{{\mathcal{V}}}
\newcommand{\calW}{{\mathcal{W}}}
\newcommand{\calX}{{\mathcal{X}}}
\newcommand{\calY}{{\mathcal{Y}}}
\newcommand{\calZ}{{\mathcal{Z}}}
\newcommand{\calbX}{\mbox{\boldmath $\mathcal{X}$}}
\newcommand{\calbY}{\mbox{\boldmath $\mathcal{Y}$}}

\newcommand{\bcalA}{\mbox{\boldmath $\calA$}}
\newcommand{\bcalB}{\mbox{\boldmath $\calB$}}
\newcommand{\bcalC}{\mbox{\boldmath $\calC$}}
\newcommand{\bcalD}{\mbox{\boldmath $\calD$}}
\newcommand{\bcalE}{\mbox{\boldmath $\calE$}}
\newcommand{\bcalF}{\mbox{\boldmath $\calF$}}
\newcommand{\bcalG}{\mbox{\boldmath $\calG$}}
\newcommand{\bcalH}{\mbox{\boldmath $\calH$}}
\newcommand{\bcalI}{\mbox{\boldmath $\calI$}}
\newcommand{\bcalJ}{\mbox{\boldmath $\calJ$}}
\newcommand{\bcalK}{\mbox{\boldmath $\calK$}}
\newcommand{\bcalL}{\mbox{\boldmath $\calL$}}
\newcommand{\bcalM}{\mbox{\boldmath $\calM$}}
\newcommand{\bcalN}{\mbox{\boldmath $\calN$}}
\newcommand{\bcalO}{\mbox{\boldmath $\calO$}}
\newcommand{\bcalP}{\mbox{\boldmath $\calP$}}
\newcommand{\bcalQ}{\mbox{\boldmath $\calQ$}}
\newcommand{\bcalR}{\mbox{\boldmath $\calR$}}
\newcommand{\bcalS}{\mbox{\boldmath $\calS$}}
\newcommand{\bcalT}{\mbox{\boldmath $\calT$}}
\newcommand{\bcalU}{\mbox{\boldmath $\calU$}}
\newcommand{\bcalV}{\mbox{\boldmath $\calV$}}
\newcommand{\bcalW}{\mbox{\boldmath $\calW$}}
\newcommand{\bcalX}{\mbox{\boldmath $\calX$}}
\newcommand{\bcalY}{\mbox{\boldmath $\calY$}}
\newcommand{\bcalZ}{\mbox{\boldmath $\calZ$}}

\newcommand{\sfA}{\mbox{$\mathsf A$}}
\newcommand{\sfB}{\mbox{$\mathsf B$}}
\newcommand{\sfC}{\mbox{$\mathsf C$}}
\newcommand{\sfD}{\mbox{$\mathsf D$}}
\newcommand{\sfE}{\mbox{$\mathsf E$}}
\newcommand{\sfF}{\mbox{$\mathsf F$}}
\newcommand{\sfG}{\mbox{$\mathsf G$}}
\newcommand{\sfH}{\mbox{$\mathsf H$}}
\newcommand{\sfI}{\mbox{$\mathsf I$}}
\newcommand{\sfJ}{\mbox{$\mathsf J$}}
\newcommand{\sfK}{\mbox{$\mathsf K$}}
\newcommand{\sfL}{\mbox{$\mathsf L$}}
\newcommand{\sfM}{\mbox{$\mathsf M$}}
\newcommand{\sfN}{\mbox{$\mathsf N$}}
\newcommand{\sfO}{\mbox{$\mathsf O$}}
\newcommand{\sfP}{\mbox{$\mathsf P$}}
\newcommand{\sfQ}{\mbox{$\mathsf Q$}}
\newcommand{\sfR}{\mbox{$\mathsf R$}}
\newcommand{\sfS}{\mbox{$\mathsf S$}}
\newcommand{\sfT}{\mbox{$\mathsf T$}}
\newcommand{\sfU}{\mbox{$\mathsf U$}}
\newcommand{\sfV}{\mbox{$\mathsf V$}}
\newcommand{\sfW}{\mbox{$\mathsf W$}}
\newcommand{\sfX}{\mbox{$\mathsf X$}}
\newcommand{\sfY}{\mbox{$\mathsf Y$}}
\newcommand{\sfZ}{\mbox{$\mathsf Z$}}

\newcommand{\balpha}{\mbox{\boldmath $\alpha$}}
\newcommand{\bbeta}{\mbox{\boldmath $\beta$}}
\newcommand{\bgamma}{\mbox{\boldmath $\gamma$}}
\newcommand{\bdelta}{\mbox{\boldmath $\delta$}}
\newcommand{\bepsilon}{\mbox{\boldmath $\epsilon$}}
\newcommand{\bvarepsilon}{\mbox{\boldmath $\varepsilon$}}
\newcommand{\bzeta}{\mbox{\boldmath $\zeta$}}
\newcommand{\boldeta}{\mbox{\boldmath $\eta$}}
\newcommand{\btheta}{\mbox{\boldmath $\theta$}}
\newcommand{\bvartheta}{\mbox{\boldmath $\vartheta$}}
\newcommand{\biota}{\mbox{\boldmath $\iota$}}
\newcommand{\bkappa}{\mbox{\boldmath $\kappa$}}
\newcommand{\blambda}{\mbox{\boldmath $\lambda$}}
\newcommand{\bmu}{\mbox{\boldmath $\mu$}}
\newcommand{\bnu}{\mbox{\boldmath $\nu$}}
\newcommand{\bxi}{\mbox{\boldmath $\xi$}}
\newcommand{\bpi}{\mbox{\boldmath $\pi$}}
\newcommand{\bvarpi}{\mbox{\boldmath $\varpi$}}
\newcommand{\brho}{\mbox{\boldmath $\rho$}}
\newcommand{\bvarrho}{\mbox{\boldmath $\varrho$}}
\newcommand{\bsigma}{\mbox{\boldmath $\sigma$}}
\newcommand{\bvarsigma}{\mbox{\boldmath $\varsigma$}}
\newcommand{\btau}{\mbox{\boldmath $\tau$}}
\newcommand{\bupsilon}{\mbox{\boldmath $\upsilon$}}
\newcommand{\bphi}{\mbox{\boldmath $\phi$}}
\newcommand{\bvarphi}{\mbox{\boldmath $\varphi$}}
\newcommand{\bchi}{\mbox{\boldmath $\chi$}}
\newcommand{\bpsi}{\mbox{\boldmath $\psi$}}
\newcommand{\bomega}{\mbox{\boldmath $\omega$}}

\newcommand{\bGamma}{\mbox{\boldmath $\Gamma$}}
\newcommand{\bDelta}{\mbox{\boldmath $\Delta$}}
\newcommand{\bTheta}{\mbox{\boldmath $\Theta$}}
\newcommand{\bLambda}{\mbox{\boldmath $\Lambda$}}
\newcommand{\bXi}{\mbox{\boldmath $\Xi$}}
\newcommand{\bPi}{\mbox{\boldmath $\Pi$}}
\newcommand{\bSigma}{\mbox{\boldmath $\Sigma$}}
\newcommand{\bUpsilon}{\mbox{\boldmath $\Upsilon$}}
\newcommand{\bPhi}{\mbox{\boldmath $\Phi$}}
\newcommand{\bPsi}{\mbox{\boldmath $\Psi$}}
\newcommand{\bOmega}{\mbox{\boldmath $\Omega$}}

\newcommand{\veca}{{\vec{\ba}}}
\newcommand{\vecb}{{\vec{\bb}}}
\newcommand{\vecc}{{\vec{\bc}}}
\newcommand{\vecd}{{\vec{\bd}}}
\newcommand{\vece}{{\vec{\bolde}}}
\newcommand{\vecf}{{\vec{\boldf}}}
\newcommand{\vecg}{{\vec{\bg}}}
\newcommand{\vech}{{\vec{\bh}}}
\newcommand{\veci}{{\vec{\bi}}}
\newcommand{\vecj}{{\vec{\bj}}}
\newcommand{\veck}{{\vec{\bk}}}
\newcommand{\vecl}{{\vec{\bl}}}
\newcommand{\vecm}{{\vec{\bm}}}
\newcommand{\vecn}{{\vec{\bn}}}
\newcommand{\veco}{{\vec{\bo}}}
\newcommand{\vecp}{{\vec{\bp}}}
\newcommand{\vecq}{{\vec{\bq}}}
\newcommand{\vecr}{{\vec{\br}}}
\newcommand{\vecs}{{\vec{\bs}}}
\newcommand{\vect}{{\vec{\bt}}}
\newcommand{\vecu}{{\vec{\bu}}}
\newcommand{\vecv}{{\vec{\bv}}}
\newcommand{\vecw}{{\vec{\bw}}}
\newcommand{\vecx}{{\vec{\bx}}}
\newcommand{\vecy}{{\vec{\by}}}
\newcommand{\vecz}{{\vec{\bz}}}

\newcommand{\vecxi}{{\vec{\bxi}}}
\newcommand{\vecphi}{{\vec{\bphi}}}
\newcommand{\vecvarphi}{{\vec{\bvarphi}}}
\newcommand{\vecbeta}{{\vec{\bbeta}}}
\newcommand{\vecdelta}{{\vec{\bdelta}}}
\newcommand{\vectheta}{{\vec{\btheta}}}

\newcommand{\Real}{\mathbb R}
\newcommand{\Complex}{\mathbb C}
\newcommand{\Natural}{\mathbb N}
\newcommand{\Integer}{\mathbb Z}


\newcommand{\bone}{\mbox{\boldmath $1$}}
\newcommand{\bzero}{\mbox{\boldmath $0$}}
\newcommand{\0}{{\bf 0}}

\newcommand{\be}{\begin{eqnarray}}
\newcommand{\ee}{\end{eqnarray}}
\newcommand{\bee}{\begin{eqnarray*}}
\newcommand{\eee}{\end{eqnarray*}}

\newcommand{\matrixb}{\left[ \begin{array}}
\newcommand{\matrixe}{\end{array} \right]}

\newcommand{\argmax}{\operatornamewithlimits{\arg \max}}
\newcommand{\argmin}{\operatornamewithlimits{\arg \min}}

\newcommand{\mean}[1]{\left \langle #1 \right \rangle}
\newcommand{\ave}{\mathbb E}
\newcommand{\E}{\mathbb E}
\newcommand{\empha}[1]{{\color{red} \bf #1}}
\newcommand{\fracpartial}[2]{\frac{\partial #1}{\partial  #2}}
\newcommand{\incomplete}[1]{\textcolor{red}{#1}}

\def\doublespace{\renewcommand{\baselinestretch}{2}\large\normalsize}
\def\singlespace{\renewcommand{\baselinestretch}{1}\large\normalsize}
\def\onehalfspace{\renewcommand{\baselinestretch}{1.5}\large\normalsize}
\def\onequaterspace{\renewcommand{\baselinestretch}{1.3}\large\normalsize}
\def\threequaterspace{\renewcommand{\baselinestretch}{1.7}\large\normalsize}
\def\smallspace{\renewcommand{\baselinestretch}{-.9}\large\normalsize}
\def\tinyspace{\renewcommand{\baselinestretch}{-.7}\large\normalsize}

\newcommand{\tr} { \textrm{tr} }
\newcommand{\re} { \textrm{re} }
\newcommand{\im} { \textrm{im} }
\newcommand{\diag} { \textrm{diag} }
\newcommand{\ddiag} { \textrm{ddiag} }
\newcommand{\off} { \textrm{off} }
\newcommand{\vectxt} { \textrm{vec} }

\newcommand{\lla}{\left\langle}
\newcommand{\rra}{\right\rangle}
\newcommand{\llbr}{\left\lbrack}
\newcommand{\rrbr}{\right\rbrack}
\newcommand{\llb}{\left\lbrace}
\newcommand{\rrb}{\right\rbrace}


\newcommand{\RR}{I\!\!R} 
\newcommand{\Nat}{I\!\!N} 
\newcommand{\CC}{I\!\!\!\!C} 

\newcommand{\Tref}[1]{Table~\ref{#1}}
\newcommand{\Eref}[1]{Eq.~(\ref{#1})}
\newcommand{\Fref}[1]{Fig.~\ref{#1}}
\newcommand{\FCref}[1]{Chapter.~\ref{#1}}
\newcommand{\Sref}[1]{Sec.~\ref{#1}}
\newcommand{\Aref}[1]{Algo.~\ref{#1}}

\def\eg{\emph{e.g.}}
\def\Eg{\emph{E.g. }}
\def\etal{\emph{et al. }}
\def\ie{\emph{i.e. }}

\title{Long-range Multimodal Pretraining for Movie Understanding}

\author{Dawit Mureja Argaw \\ KAIST \and Joon-Young Lee \\ Adobe \and Markus Woodson \\ Adobe \and In So Kweon \\ KAIST \and Fabian Caba Heilbron \\ Adobe}

\maketitle

\begin{abstract}
Learning computer vision models from (and for) movies has a long-standing history. While great progress has been attained, there is still a need for a pretrained multimodal model that can perform well in the ever-growing set of movie understanding tasks the community has been establishing. In this work, we introduce Long-range Multimodal Pretraining, a strategy, and a model that leverages movie data to train transferable multimodal and cross-modal encoders. Our key idea is to learn from all modalities in a movie by observing and extracting relationships over a long-range. After pretraining, we run ablation studies on the LVU benchmark and validate our modeling choices and the importance of learning from long-range time spans. Our model achieves state-of-the-art on several LVU tasks while being much more data efficient than previous works. Finally, we evaluate our model's transferability by setting a new state-of-the-art in five different benchmarks.
\end{abstract}

\section{Introduction}
Are movies just for entertainment? Arguably they offer much more than that. Movies are a source of inspiration for many and a force that influences societal behaviors~\cite{nyt_article}. Movies are also an active topic of study in the computer vision community~\cite{laptev2008learning, everingham2006hello, vicol2018moviegraphs, bain2020condensed, huang2020movienet, argaw2022anatomy}. They have served as a testbed for measuring progress in visual recognition \cite{gu2018ava, bose2022movieclip}, reasoning \cite{sadhu2021visual, tapaswi2016movieqa}, and creative editing \cite{courant2021high, pardo2021moviecuts}. Moreover, movie data has been also leveraged to train computer vision \cite{chen2022movies2scenes}, machine listening \cite{chaudhuri2018ava}, and NLP \cite{zheng2020pre} systems. Besides entertainment, movies are surely an intriguing media that can help AI models to understand semantics and artistic expressions encoded in a movie style. 

While the quest of understanding movies has gained steady attention, it is still an open question how to develop models that leverage abundant sources of movie data \cite{huang2020movienet, soldan2022mad} to tackle all the movie-related tasks that the community has grown over the last decade~\cite{bain2020condensed, liu2021multi, wu2021towards, suris2022s, argaw2022anatomy}. But, learning from movies is not a straightforward task, especially when no labels are available.

Existing video self-supervised approaches~\cite{dave2022tclr, feichtenhofer2022masked, akbari2021vatt, xu2021videoclip, zellers2022merlot}, which primarily focus on learning from short clips, would not leverage the richness of movies, as the value of movies as training sources emerges from their long-range dependencies. Moreover, the end-to-end learning scheme adopted in these works can not be easily extended to movies as it is computationally infeasible to encode long-form sequences in an end-to-end manner.

\begin{figure}[!t]
    \centering
    \includegraphics[width=1.0\linewidth]{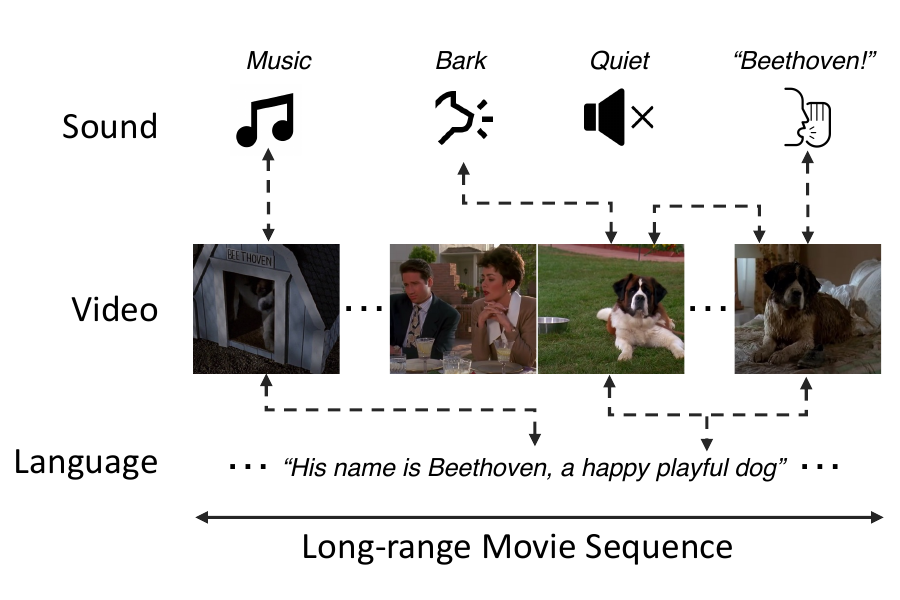}
    \caption{\textbf{Learning from Movie Sequences.} Movie sequences offer learning signals when observed for long-ranges. By reading the language (dialogue), hearing the sounds, and looking at the video, one can determine that Beethoven, the dog, was barking and he sleeps in a white and blue wooden dog house.}
    \label{fig:teaser}
\end{figure}

Recent works on long-form video understanding~\cite{wu2021towards, islam2022long, chen2022movies2scenes} have attempted to address these limitations by encoding movie clips using frozen base encoders~\cite{wu2021towards, chen2022movies2scenes} or state-space models~\cite{islam2022long}. However, these works exclusively focus on the video modality for long-range temporal reasoning, disregarding audio and text signals, and hence are limited to specific tasks. In this work, we argue that long-form videos such as movies are rich in visual, auditory, and textual information, and integrating these modalities would lead to a better and generalizable understanding.

\Fref{fig:teaser} illustrates how the three core modalities in movies (language, video, sound) can serve as a valuable source of supervision when reasoned all together for a long time. By analyzing the language (from the dialogue) we can understand that the dog's name is Beethoven, from the audio we could hear he is barking, and from long-range associations, we can infer he sleeps in a wooden dog house. This toy example illustrates how critical is to design training strategies that can effectively encode visual, audio, and language relationships. It is equally important to effectively design models that encode long-range dependencies.

This work presents pretraining strategy that leverages multimodal cues and long-range reasoning in movies. We develop a flexible model that can be easily transferred to a range of tasks related to movie analysis and understanding~\cite{wu2021towards, liu2021multi, argaw2022anatomy, suris2022s, bain2020condensed}. Our \textit{first} design requirement is that the model needs to observe over a long time span in order to do long-range reasoning. To facilitate this, we dissect a long video into shots and represent it as an ordered sequence of shots. The main motivation for using shot-based input sampling instead of uniform temporal sampling is to capture longer multi-shot content at a time since each shot will be considered as one token irrespective of its length. Our \textit{second} design requirement is to efficiently encode the input sequence by harnessing all available modalities. We do so by representing each shot clip in the sequence in video, audio, and language modalities.  As it is computationally inefficient to encode long-form sequences in an end-to-end manner, we opt for using pretrained state-of-the-art models~\cite{tran2018closer, hershey2017cnn, baevski2020wav2vec, lewis2019bart} as base encoders to transform the raw video, audio, and text sequences into their corresponding compact feature representations. 

Given a sequence of encoded base features for each modality, our \textit{third} design requirement is to perform multimodal long-term reasoning in a self-supervised manner. To do so, we make use of Transformer networks~\cite{vaswani2017attention}. Instead of combining all tokens from the different modalities as one long sequence like in VideoBERT~\cite{sun2019videobert}, we follow a \textit{hierarchical} approach. We first learn long-range context from each modality using \textit{contextual} transformers while simultaneously ensuring that the context learned over one modality is also conditioned by another modality. We then learn joint representations between modalities using a \textit{cross-modal} transformer network. To ensure that different transformers in our framework serve their purpose, we introduce a pretraining strategy that enforces intra-modal, inter-modal, and cross-modal relationships over long-range observations via carefully designed losses. 

We train our model using publicly available movie dataset~\cite{soldan2022mad}, performing additional preprocessing such as shot boundary detection~\cite{souvcek2020transnet}. We evaluate the transferability of our approach on six different benchmarks~\cite{wu2021towards, liu2021multi, argaw2022anatomy, suris2022s, bain2020condensed}, and empirically show that long-range multimodal pretraining provides extensive benefits in performance.
\vspace{-3mm}
\paragraph{Contributions.} Our goal is to train transferable models for movie understanding. It brings two contributions:\\
\textbf{(1)}  We introduce a pretraining strategy designed to leverage long-range multimodal cues in movies. We propose a model that captures intra-modal, inter-modal, and cross-modal dependencies via transformer encoders and self-supervision.\\
\textbf{(2)}  We conduct extensive experiments to validate the transferability of our model and the contributions of the pretraining strategy. The results show that our model consistently improves the state-of-the-art across six benchmarks. 
\section{Related Works}
\paragraph{Multimodal Pretraining in Long Videos.} Pretraining from long videos has recently gained lots of attention \cite{zellers2022merlot, sun2022long, akbari2021vatt, xu2021videoclip}. Previous works have built upon the intuition that modeling multimodality in long videos is key for learning. For instance, Zellers \etal \cite{zellers2022merlot} introduced an end-to-end BERT-alike model \cite{devlin2018bert} that learns from sound, video, and transcripts from narrated YouTube videos. Similarly, \cite{akbari2021vatt} introduced multimodal models that use contrastive objectives to train base encoders on a large-scale collection of videos. While these approaches can offer specialized encoders for general perception tasks, their design choice for training the base encoders end-to-end prevents them to encode longer sequences and effectively modeling long-range dependencies. Concurrent to our work, Sun \etal introduced a framework for video-language pretraining in long videos \cite{sun2022long}. Their proposed method shares some design choices with ours, such as including both video-language contrastive objectives, but also a cross-modal module trained with mask language modeling objectives \cite{devlin2018bert}. Our method differs from these approaches in three key aspects: (i) we opt for leveraging frozen base encoders to facilitate longer modeling; (ii) our cross-modal objective is anchored on reconstructing the video representations from both audio and language modalities; and (iii) our method is specialized to movies and is pretrained using a much smaller dataset.

\paragraph{Learning from Movies.} Movies have served as a popular source for training computer vision models \cite{laptev2008learning, moviegraphs, rao2020unified, xu2011using}. They have facilitated the creation of multiple datasets, enabling early research in action recognition~\cite{laptev2008learning}, substantial progress in character recognition~\cite{everingham2006hello}, and studies in cinematography~\cite{xu2011using,argaw2022anatomy,chen2022match}, among many other areas~\cite{zhu2015aligning,bain2020condensed,suris2022s}. Besides, movies have been also leveraged to train self-supervised models for various applications \cite{pardo2021learning,pavlakos2022multishot,pavlakos2022one}. Finally, and closer to ours, recent methods have harnessed movies to train video representations for movies \cite{kalayeh2021watching,xiao2022hierarchical,islam2022long,chen2022movies2scenes}. Different from our approach, these works focus on training unimodal (visual) representations \cite{xiao2022hierarchical,chen2022movies2scenes,islam2022long} or do not model long-range sequences \cite{kalayeh2021watching}.
\begin{figure*}[!t]
    \centering
    \includegraphics[width=1.0\linewidth,trim={0.75cm 2.0cm 4.5cm 2.3cm},clip]{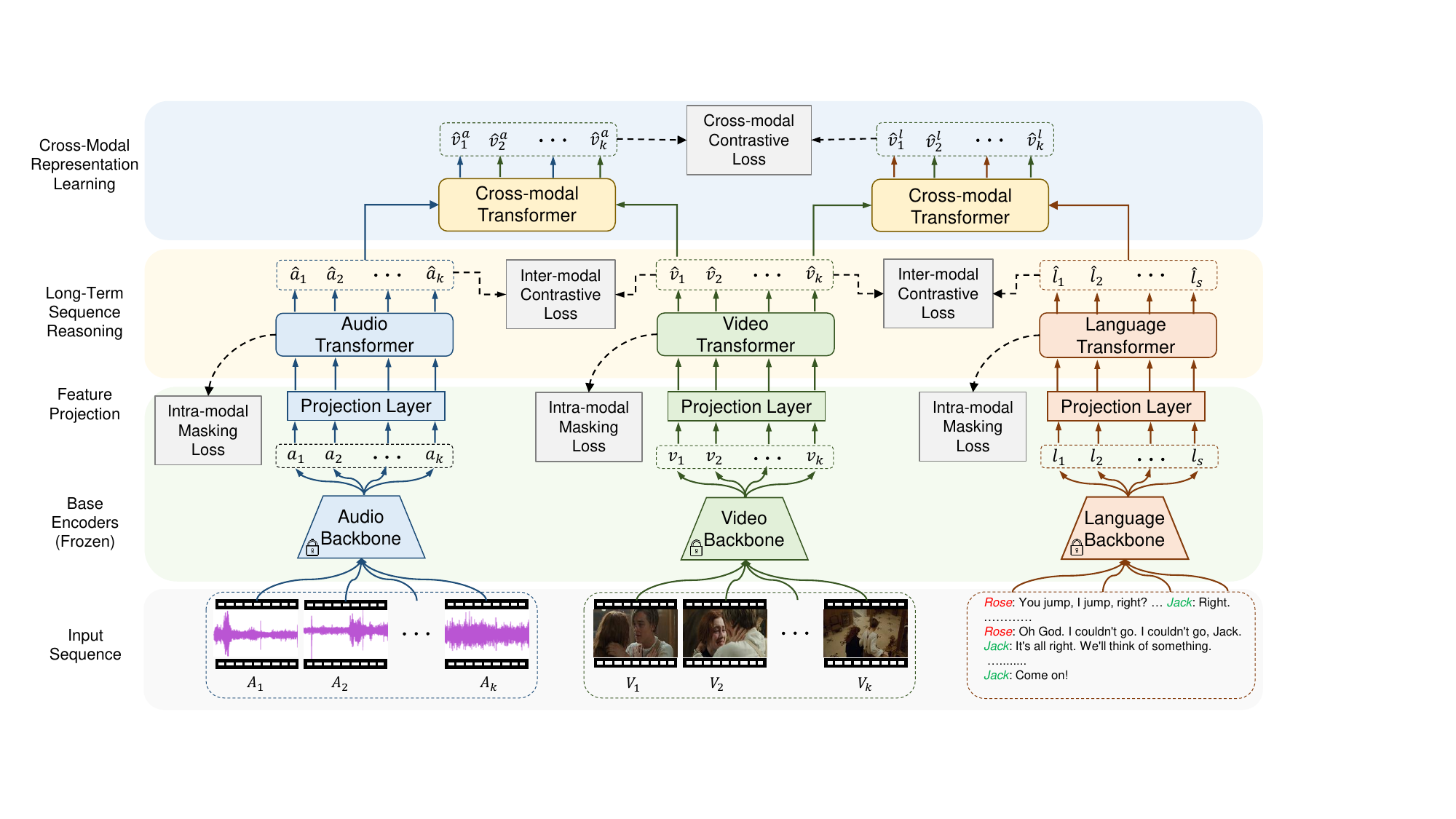}
    \caption{\textbf{Long-range Multimodal Pretraining.} Our approach takes as input audio, video, and language observations extracted from a sequence of $k$ consecutive movie shots. The observations from each modality are encoded with \textit{frozen} base encoders, and its outputs are projected via projection layers. Then, a stack of transformer encoders contextualize the base features of each modality. Finally, a cross-modal transformer is used to learn joint embeddings between the contextualized features. Our pretraining includes intra-modal, inter-modal and cross-modal losses to jointly train the transformer models. The pretrained model can be used as a backbone encoder for several downstream tasks.}
    \label{fig:overview}
\end{figure*}
\paragraph{Movie Understanding.} There has been a growing interest in understanding long-form videos such as movies and TV shows with AI. To facilitate research in this direction, several datasets and benchmarks have been proposed. Huang~\etal\cite{huang2020movienet} proposed MovieNet, a dataset that contains 1,100 full movies and 60K trailers with annotations for various tasks including scene segmentation, character recognition, and genre prediction. Wu~\etal\cite{wu2021towards} introduced the LVU benchmark which contains a total of 9 tasks related to content understanding, metadata prediction, and user engagement. Liu~\etal\cite{liu2021multi} collected a large-scale dataset from TV shows for multi-shot temporal event localization. Some other works proposed~\cite{bain2020condensed,suris2022s} movie clips-based datasets for text-video and audio-video retrieval tasks. Recently, \cite{argaw2022anatomy} introduced the AVE benchmark for AI-assisted video editing. Our work offers a transferable model that can be tasked to address various movie understanding tasks.
\section{Long-range Multimodal Pretraining}

Given a movie $M$, we first dissect $M$ into a sequence of shots using a shot-boundary detector~\cite{souvcek2020transnet}. Then, we sample $k$ consecutive shots~\ie~$\{S_{1}, S_{2},\ldots, S_{k}\}$, as an input to our method. Our motivation for such a design choice comes in twofold. First, shot-based sampling captures longer multi-shot content at a time compared to uniform temporal sampling, since each shot is considered as one token irrespective of its length. Second, a model trained in this manner could be easily deployed to video editing tasks~\cite{argaw2022anatomy}, where shots from different camera setups are sequentially assembled to make an edited scene. Each shot clip in the input sequence is represented in video ($V$), audio ($A$), and language ($L$) modalities as shown in \Fref{fig:overview}. The video and audio inputs are directly extracted from the given shot sequence. For language input, we use all the text data spanning between the first and the last shot in the sequence.

\subsection{Base Feature Encoding}
\label{sec:meth_part1}

As it is computationally inefficient to encode long sequences in an end-to-end manner, we opt for using state-of-the-art pretrained networks as base encoders. For the video inputs, $\{V_{1}, V_{2},\ldots, V_{k}\}$, the corresponding base video features,~\ie~$\{v_{1}, v_{2},\ldots, v_{k}\}$, are obtained by passing each shot clip in the sequence into a frozen video backbone. In the same manner, base audio features, \ie~$\{a_{1}, a_{2},\ldots, a_{k}\}$, are encoded from the input audio sequence using an audio backbone network. The base language features, \ie~$\{l_{1}, l_{2},\ldots, l_{s}\}$ where $s$ denotes the total number of word tokens, are extracted by encoding the raw text data using a pretrained language model. The goal of this step is to capture local information from each source and compress the captured information into a set of tokens to be used for the next steps. After base feature encoding, we project the encoded tokens across different modalities into a matching dimension using a linear layer.

\subsection{Long-term Sequence Reasoning}
\label{sec:meth_part2}

Given a sequence of base features, we aim to design a model that integrates all available modalities in a self-supervised manner to perform long-term reasoning. First, we feed the base feature sequence from each modality into Transformer encoders~\cite{vaswani2017attention} as shown in \Fref{fig:overview}. The purpose of these transformers is to learn temporal context from the long-range input as each element in the given sequence attends to every other element in the sequence. Thus, we refer to them as \textit{contextual} transformers. To ensure that the contextual transformers serve their purpose, we impose the following two constraints. First, we adopt a BERT-like~\cite{devlin2018bert} pretraining, where we replace 20\% of the tokens with a special \textsf{\footnotesize MASK} token for audio, video, and text sequences. Each transformer is then trained to match the \textsf{\footnotesize MASK}ed prediction with the corresponding input representation via intra-modality masking loss. For simplicity, let $\hat{w}$ denote the predicted representation for a masked input token $w$ in a batch of $\calW$ tokens. The \textit{intra-modal} masking loss is then formulated as minimizing the cross entropy between $\hat{w}$ and $w$ in all modalities as shown in \Eref{eqn:cross_ent} and \Eref{eqn:intra_modal}.
\begin{equation}
\small
    \calL_\mathrm{mask} (\cdot) = \frac{1}{|\calW|} \sum_{w\in \calW} \Big(\log \frac{\exp(\hat{w} \cdot w)}{\sum_{w\in \calW} \exp(\hat{w} \cdot w)}\Big)
    \label{eqn:cross_ent}
\end{equation}
\begin{equation}
\small
    \calL_\mathrm{intra-modal} = \calL_\mathrm{mask} (A) + \calL_\mathrm{mask} (V) +\calL_\mathrm{mask} (L)
    \label{eqn:intra_modal}
\end{equation}
While it is crucial to learn long-range contexts in each modality, it is equally important to ensure that the context learned over one modality is also conditioned by another modality in order to effectively capture the underlying multimodal cues in long-form videos. For instance, the soundtrack used in a movie often matches the visual content (\eg~pace, scene) of the movie. To enforce this notion in our model, we impose an inter-modal alignment between learned representations from the contextual transformers,~\ie~contextual features. Using video modality as an anchor, we define the \textit{inter-modal} contrastive loss as the sum of the InfoNCE loss~\cite{oord2018representation} between audio $\leftrightarrow$ video, and video $\leftrightarrow$ language representations. Given the one-to-one alignment between contextual audio and video features ($\{\hat{a}_1,\hat{a}_2,\ldots,\hat{a}_k\}$ and $\{\hat{v}_1,\hat{v}_2,\ldots,\hat{v}_k\}$), we encourage the corresponding (positive) pairs to have high similarity and non-corresponding (negative) pairs to have low similarity as formulated in \Eref{eqn:loss_v_a}.

\begin{equation}
\small
    \calL_{\hat{v} \rightarrow \hat{a}} = -\sum_{j}^{N} \sum_{i}^{k}\Bigg(\log \frac{\exp (\calS(\hat{v}_{j,i}, \hat{a}_{j,i}) / \tau)}{\sum_{j}^{N} \sum_{i}^{k} \exp\big(\calS(\hat{v}_{j,i}, \hat{a}_{j,i}) / \tau\big)}\Bigg)
    \label{eqn:loss_v_a}
\end{equation}
, where $j$ and $i$ index batch size $N$ and input sequence length $k$, respectively, $\calS$ denotes cosine similarity and $\tau$ is a temperature parameter. The contrastive loss between contextual video and language features ($\{\hat{v}_1,\hat{v}_2,\ldots,\hat{v}_k\}$ and $\{\hat{l}_1,\hat{l}_2,\ldots,\hat{l}_s\}$) is implemented using a one-to-many alignment scheme, \ie~for each video feature $\hat{v}_i$, the corresponding pair is obtained by averaging the representations of the language tokens spanning the interval of $\hat{v}_i$ such that:
\begin{equation}
\small
    \Bar{l}_i = \frac{1}{C_i}\sum_{c = 1}^{C_i} \hat{l}_c
    \label{eqn:l_comp}
\end{equation}
, where $C_i$ represents the number of language tokens in the interval of shot $S_i$. This is equivalent to matching each shot $S_i$ to its corresponding textual representation. The resulting loss is then formulated similarly to the loss in \Eref{eqn:loss_v_a} using $\hat{v}_i$ and $\Bar{l}_i$. The total inter-modal contrastive loss is computed symmetrically as follows.
\begin{equation}
\small
    \calL_\mathrm{inter-modal} = \calL_{\hat{v} \rightarrow \hat{a}} + \calL_{\hat{a} \rightarrow \hat{v}} + \calL_{\hat{v} \rightarrow \Bar{l}} + \calL_{\Bar{l} \rightarrow  \hat{v}}
    \label{eqn:inter_modal}
\end{equation}

\begin{figure*}[!t]
    \centering
    \includegraphics[width=1.0\linewidth,trim={0.5cm 3.5cm 0.9cm 6cm},clip]{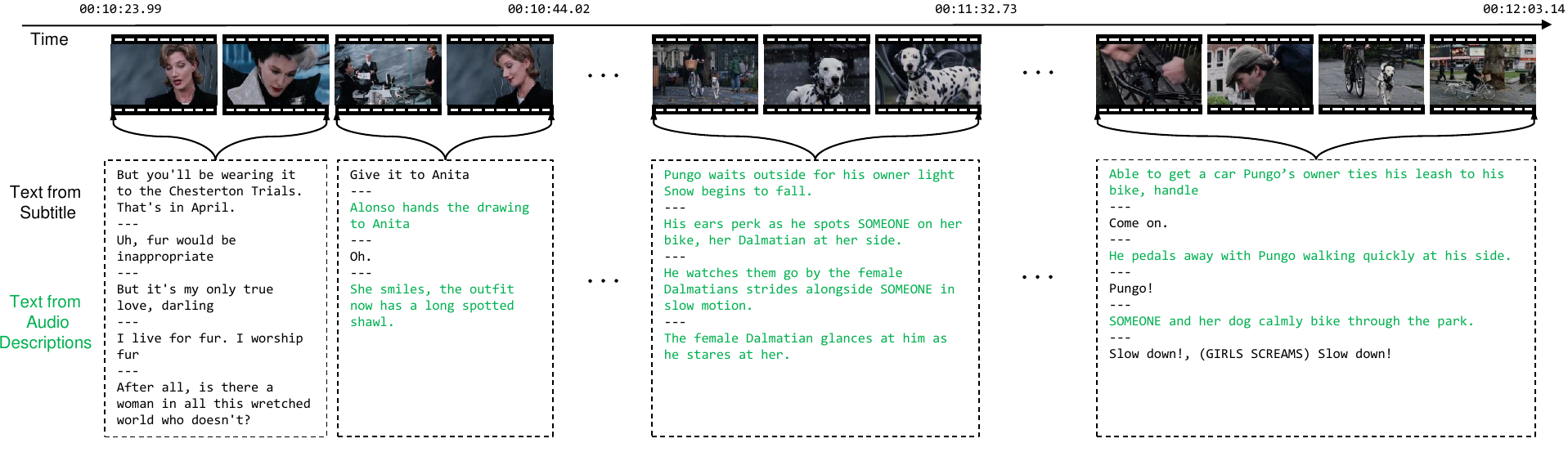}
    \caption{Example of the input sequence of shots taken from the movie `101 Dalmatians (1996)'. The corresponding raw language data is composed of temporally aligned texts from both the subtitle data (\textit{where there is speech}) and transcribed audio descriptions~\cite{soldan2022mad} (\textit{where there is no speech}).}
    \label{fig:dataset_details}
\end{figure*}

\subsection{Cross-modal Representation Learning}
\label{sec:meth_part3}

Thus far, our model learns to reason over long-term sequences in different modalities while simultaneously exploring their correlation. Intuitively speaking, the losses defined in \Eref{eqn:intra_modal} and \Eref{eqn:inter_modal} should be strong enough constraints to learn transferable representations. However, the interaction between the different modalities is enforced only at a loss level so far. Inspired by the success of joint encoding in related works~\cite{sun2019videobert, akbari2021vatt}, we adopt a cross-modal Transformer~\cite{vaswani2017attention} network to explicitly facilitate joint representation learning in our framework. The key motivation here is to further exploit multimodal cues from audio and language features to guide long-form video representation learning. We accomplish this by feeding the contextual audio/language features as a \textit{source} sequence into the encoder, and the contextual video features as a \textit{target} sequence into the decoder of the cross-modal transformer as shown in~\Fref{fig:overview}. Thus, our model decodes audio-conditioned and language-conditioned visual features, \ie~$\{\hat{v}^a_1,\hat{v}^a_2,\ldots,\hat{v}^a_k\}$ and $\{\hat{v}^l_1,\hat{v}^l_2,\ldots,\hat{v}^l_k\}$, by encoding the audio and language representations, respectively. To ensure that the cross-modal transformer serves its purpose, we introduce a cross-modal contrastive loss that encourages alignment between the learned cross-modal representations. It is defined as the InfoNCE loss~\cite{oord2018representation} between audio-conditioned ($\{\hat{v}^a_1,\hat{v}^a_2,\ldots,\hat{v}^a_k\}$) and language conditioned ($\{\hat{v}^l_1,\hat{v}^l_2,\ldots,\hat{v}^l_k\}$) visual features as shown in \Eref{eqn:cross_modal}.

\begin{equation}
    \calL_\mathrm{cross-modal} = \calL_{\hat{v}^a \rightarrow \hat{v}^l} + \calL_{\hat{v}^l \rightarrow \hat{v}^a }
    \label{eqn:cross_modal}
\end{equation}

While the base feature encoders stay frozen, the remaining modules in the proposed framework are end-to-end trained by jointly optimizing the losses in \Eref{eqn:intra_modal}, \Eref{eqn:inter_modal} and \Eref{eqn:cross_modal}. Therefore, the total pretraining loss to self-supervise our model is defined by:
\begin{equation}
\small
    \calL_\mathrm{total} = \calL_\mathrm{intra-modal} + \calL_\mathrm{inter-modal} + \calL_\mathrm{cross-modal}
        \label{eqn:loss_total}
\end{equation}

\paragraph{Pretraining Dataset.} We use the Movie Audio Description (MAD)~\cite{soldan2022mad} dataset to train our model. The dataset contains a diverse set of movies making more than 1200 hours of content. To ensure that every segment in a given movie has a corresponding language modality\footnote{MAD provides textual data only for movie segments with no speech.}, we collect the official subtitle data for each movie and temporally align them with the textual descriptions from MAD as shown in \Fref{fig:dataset_details}. We prepare our training data by extracting video, audio, and language modalities from each full-length movie. For \textit{video} modality, we use a shot-boundary detector~\cite{souvcek2020transnet} to segment a movie into a sequence of shot clips. We use FFmpeg to extract the corresponding \textit{audio} of each shot clip. For \textit{language} modality, we utilize the text data, including temporally aligned subtitles and audio descriptions, extracted from the beginning of the first shot to the end of the last shot in the sequence, while excluding overlapping content. We made sure that there is no overlap between the pretraining list and any of the videos (movies) in the test set of all downstream tasks by comparing IMDb ids.

\paragraph{Implementation Details.} We set the shot sequence length $k=30$ during pretraining. In other words, we use a 90-second movie clip as an input, since the shots in the pretraining data are approximately 3 seconds long on average. We use R(2+1)D-ResNet50~\cite{tran2018closer} model as the base video encoder. For the audio signals, we concatenate features from VGGish~\cite{hershey2017cnn} and wav2vec 2.0~\cite{baevski2020wav2vec} networks; these features encode soundtrack and speech content representations, respectively. For the language input, we use the BART~\cite{lewis2019bart} model to encode the raw text data to base language features. We adopt a 3-layer Transformer architecture~\cite{vaswani2017attention} for all contextual and cross-modal transformers. We use AdamW~\cite{loshchilov2017decoupled} optimizer, a cosine learning rate annealing strategy~\cite{touvron2021training} with an initial value of $1e-3$ and a batch size of 1024 during pretraining. For all the InfoNCE~\cite{oord2018representation} losses defined in \Eref{eqn:inter_modal} and \Eref{eqn:cross_modal} and we use a temperature parameter $\tau = 0.3$.

\section{Experiment}
\label{sec:experiment}

In this section, we first present ablations (\Sref{sec:exp_part1}) and experimental analyses (\Sref{sec:exp_part2}) on the widely used long-form video understanding (LVU) benchmark~\cite{wu2021towards}. We then show the versatility of our approach (\Sref{sec:exp_part3}) in other five movie benchmarks related to event localization~\cite{liu2021multi}, scene understanding~\cite{argaw2022anatomy}, editing pattern prediction~\cite{argaw2022anatomy}, soundtrack selection~\cite{suris2022s} and scene description retrieval~\cite{bain2020condensed}. 

\paragraph{LVU Benchmark.} The long-form video understanding (LVU) benchmark~\cite{wu2021towards} contains 9 tasks that cover various aspects of long-form videos, including content understanding (\textit{character relationship}, \textit{speaking style}, \textit{scene/place}), metadata prediction (\textit{director}, \textit{genre}, \textit{writer}, \textit{release year}), and user engagement regression (\textit{like ratio}, \textit{views}). It contains $\sim$11K videos, where each video is typically one-to-three minutes long. We set up the LVU tasks as follows. First, we use our pretrained model as an encoder to extract multimodal features from a given video. Second, we time-average the extracted sequence of features in order to get an aggregated representation for the input video. Third, we train a \textit{single} classifier/regression layer on top of the time-averaged feature for the different tasks in LVU. 

\subsection{Ablation Studies}
\label{sec:exp_part1}
In \Tref{tbl:ablation}, we study the significance of each component in the proposed framework by pretraining our model in different settings and using the pretrained models as backbone encoders for downstream tasks. We report the average top-1 accuracy for content understanding and metadata prediction tasks, whereas average mean-squared error is used to evaluate user engagement tasks.

\paragraph{Pretraining.} To show the importance of the pretraining step, we train the proposed model directly on downstream tasks from scratch. This resulted in poor performance compared to pretrained models as can be seen from \Tref{tbl:ablation}. This is mainly because the tasks in LVU relatively have a small number of training samples which causes the model to overfit. More importantly, the classification/regression losses are not strong enough constraints to learn multimodal reasoning from long-form inputs. 

\begin{table}[!t]
\begin{center}
\setlength{\tabcolsep}{4.5pt}
\renewcommand{\arraystretch}{1}
\caption{\textbf{Ablation study} on the different losses during pretraining}
\vspace{1mm}
\mytabular{
\begin{tabular}{ccccccc}
\toprule
Pretrain & $\calL_\mathrm{intra}$ & $\calL_\mathrm{inter}$ & $\calL_\mathrm{cross}$ & Content $\uparrow$ & Metadata $\uparrow$ &  User $\downarrow$  \\ \midrule
\redxmark & \redxmark & \redxmark & \redxmark & 44.87 & 32.47& 3.19 \\
\greencheckmark & \redxmark & \greencheckmark & \greencheckmark & 60.13 & 55.92 & 1.23 \\
\greencheckmark & \greencheckmark & \redxmark & \greencheckmark & 52.74 & 44.86 & 1.82 \\
\greencheckmark & \greencheckmark & \greencheckmark & \redxmark & 59.36 & 54.71  & 1.34 \\
\greencheckmark & \greencheckmark & \greencheckmark & \greencheckmark & \textbf{60.53}  & \textbf{57.67}  & \textbf{1.16} \\
\bottomrule
\end{tabular}
}
\label{tbl:ablation}
\end{center}
\vspace{-6.5mm}
\end{table}

\paragraph{Intra-modal Masking.} Here, we analyze the benefit of the intra-modal contrastive loss ($\calL_\mathrm{intra-modal}$) for long-term sequence reasoning (\Sref{sec:meth_part2}). We do so by training our network without masking. As can be inferred \Tref{tbl:ablation}, a model pretrained without $\calL_\mathrm{intra-modal}$ still gives a good performance on downstream tasks. This is intuitive because the inter-modal contrastive loss ($\calL_\mathrm{inter-modal}$) should implicitly guide each contextual transformer to reason over its input sequence, ~\ie the model will effectively align the contextual representations only when it properly learns the context over each modality first. However, explicitly adding $\calL_\mathrm{intra-modal}$ during pretraining gave a performance boost as shown in \Tref{tbl:ablation}.

\paragraph{Inter-modal Alignment.} To examine the importance of imposing inter-modality alignment over the representations learned by contextual transformers (\Sref{sec:meth_part2}), we train our model without $\calL_\mathrm{inter-modal}$. A model pretrained in this manner gives a notably worse performance. This is most likely because the long-term sequence reasoning will be performed independently without inferring the intricacies between different modalities. This in turn affects the cross-modal representation learning (\Sref{sec:meth_part3}) as the cross-modal transformer will take a shortcut by ignoring audio and language representations. Thus, the learned representations will not be robustly transferable to downstream tasks explaining the poor performance in \Tref{tbl:ablation}. These results reaffirm that inter-modality alignment is a critical loss for long-range multimodal pretraining. 

\paragraph{Cross-modal Transformer.} We analyze the contribution of joint representation learning by training our model without the cross-modal transformer, \ie~without $\calL_\mathrm{cross-modal}$. As discussed in \Sref{sec:meth_part3}, optimizing $\calL_\mathrm{intra-modal}$ and $\calL_\mathrm{inter-modal}$ together should be a sufficient constraint to learn multimodal cues from long-form inputs. The empirical results in \Tref{tbl:ablation} also confirm this notion, where a model pretrained with $\calL_\mathrm{total}$ = $\calL_\mathrm{intra-modal}$ + $\calL_\mathrm{inter-modal}$ gives a competitive performance on downstream tasks. However, explicitly learning joint representation using the cross-modal transformer resulted in better performance.

\paragraph{Sequence Length.} In \Tref{tbl:ablation2}, we study the effect of input sequence length during pretraining. To do so, we train 3 different models by setting the sequence length $k$ to 10, 30, and 60. This translates to using 30, 90, and 180-second videos as input. As can be noticed from \Tref{tbl:ablation2}, there exists a pattern where a model pretrained with longer sequences gives better performance compared to a model pretrained with shorter sequences. This is intuitive because pretraining by observing longer sequences not only facilities the mining of more multimodal cues but also makes the model robust to various input video lengths at inference time. 

\begin{table}[!t]
\begin{center}
\setlength{\tabcolsep}{4.5pt}
\renewcommand{\arraystretch}{1}
\caption{\textbf{Ablation study} on sequence length during pretraining}
\vspace{1mm}
\mytabular{
\begin{tabular}{lccc}
\toprule
Sequence length ($k$)  & Content $\uparrow$ & Metadata $\uparrow$ &  User $\downarrow$  \\ \midrule
$k = 10$ & 59.77 & 56.45 & 1.20 \\
$k = 30$ (Baseline) & 60.53 & 57.67 & 1.16 \\
$k = 60$ & 61.36 & 58.12 & 1.14\\
\bottomrule
\end{tabular}
}
\label{tbl:ablation2}
\end{center}
\end{table}

\begin{table}[!t]
\begin{center}
\setlength{\tabcolsep}{4.5pt}
\renewcommand{\arraystretch}{1}
\caption{\textbf{Analysis} on the contribution of different features}
\vspace{1mm}
\mytabular{
\begin{tabular}{lccc}
\toprule
Features  & Content $\uparrow$ & Metadata $\uparrow$ &  User $\downarrow$  \\ \midrule
$v_\mathrm{base}$ & 48.43 & 45.40 & 2.01 \\
$v_\mathrm{base}$ + $a_\mathrm{base}$ & 51.03 & 51.20 & 1.91 \\ \midrule
$\hat{v}_\mathrm{context}$ & 56.50 & 53.68 & 1.82\\
$\hat{v}_\mathrm{context}$ + $\hat{a}_\mathrm{context}$ & 59.43 & 55.75 & 1.39\\
$\hat{v}_\mathrm{context}$ + $\hat{a}_\mathrm{context}$ + $\hat{v}^a_\mathrm{cross}$ & \textbf{60.53} & \textbf{57.67} & \textbf{1.16}\\
\bottomrule
\end{tabular}
}
\label{tbl:exp_analysis}
\vspace{-4mm}
\end{center}
\end{table}

\begin{table*}[!t]
\begin{center}
\setlength{\tabcolsep}{6pt}
\renewcommand{\arraystretch}{1}
\caption{\textbf{Comparison with state-of-the-art methods} on long-form video understanding (LVU) benchmark.}
\vspace{1mm}
\mytabular{
\begin{tabular}{lcccccccccccc}
\toprule
& \multicolumn{4}{c}{Content Understanding $\uparrow$}& \multicolumn{5}{c}{Metadata Prediction $\uparrow$} & \multicolumn{3}{c}{User Engagement $\downarrow$} \\ \cmidrule(lr){2-5} \cmidrule(lr){6-10} \cmidrule(lr){11-13}
Method &  Relationship & Way-Speaking & Scene & Average & Director & Genre & Writer & Year & Average & Like & Views & Average\\ \midrule
SlowFast R101~\cite{feichtenhofer2019slowfast} & 52.4 & 35.8 & 54.7 & 47.6 & 44.9 & 53.0 & 36.3 & \textbf{52.5}  & 46.7 & 0.386 & 3.77 & 2.08\\
VideoBERT~\cite{sun2019videobert} & 52.8 & 37.9 & 54.9 & 48.5 &47.3 & 51.9 & 38.5 & 36.1 & 43.4 & 0.320 & 4.46 & 2.39\\
CLIP~\cite{radford2021learning} & 56.1 & 36.7 & 52.9  & 48.6 & 56.2 & 50.9 & 37.8 & 46.4 & 47.8 & 0.411 & 3.85 & 2.13\\
Object Transformer~\cite{wu2021towards} & 53.1 & 39.4 & 56.9 & 49.8 & 51.2 & 54.6 & 34.5 & 39.1 & 44.8 & 0.230 & 3.55 & 1.89\\
ViS4mer~\cite{islam2022long} & 57.1 & 40.8 & 67.4  & 55.1 & 62.6 & 54.7 & 48.8 & 44.7  & 52.7 & 0.260 & 3.63 & 1.95\\
Movie2Scenes~\cite{chen2022movies2scenes} & \underline{67.7} & \textbf{44.9} & 63.8 & \underline{58.8} & \textbf{65.1}& \underline{57.5} & \textbf{56.2} & 51.8 & \underline{57.6} & \textbf{0.153} & \underline{2.46} & \underline{1.31} \\
LF-VILA~\cite{sun2022long} & 61.5 & 41.3 & \textbf{68.0} & 56.9 & - & - & - & - & - & - & - & -\\
\midrule
Ours & \textbf{69.4} & \underline{44.4} & \underline{67.8} & \textbf{60.5} & \underline{64.9} & \textbf{57.7} & \underline{55.8} & \underline{52.3} & \textbf{57.7} &\underline{0.163}& \textbf{2.15} &\textbf{ 1.16}\\
\bottomrule
\end{tabular}
}
\vspace{-4mm}
\label{tbl:lvu}
\end{center}

\end{table*}

\subsection{Experimental Analyses}
\label{sec:exp_part2}
\paragraph{Learned Representations.} Given our pretrained model as a backbone encoder, we study the contribution of the different representations (features) and their combination toward downstream task performance. \Tref{tbl:exp_analysis} summarizes the results on LVU benchmark~\cite{wu2021towards}. We combine features by simply concatenating (+) their time-averaged representation before feeding them to the classification/regression layer. As can be seen from \Tref{tbl:exp_analysis}, directly using base video ($v_\mathrm{base}$) and base audio
($a_\mathrm{base}$) features results in a subpar performance. This is mainly because the frozen backbones used to encode the base features
(\Sref{sec:meth_part1}) are specialized for capturing local information from short segments, and hence are ineffective for long-form inputs. In contrast, the video and audio features encoded from the contextual transformers ($\hat{v}_\mathrm{context}$ and $\hat{a}_\mathrm{context}$) 
give a significantly better performance as shown in \Tref{tbl:exp_analysis}. This confirms the benefit of long-term sequence reasoning discussed in \Sref{sec:meth_part2}. We experimentally observed that superior results on downstream tasks are obtained when features from both contextual and cross-modal transformers are combined. For instance, it can be inferred from \Tref{tbl:exp_analysis} that the best-performing model on the LVU benchmark is  $\hat{v}_\mathrm{context}$ + $\hat{a}_\mathrm{context}$ + $\hat{v}^a_\mathrm{cross}$, where $\hat{v}^a_\mathrm{cross}$ denotes the audio-conditioned visual feature. 

\paragraph{Comparison with State-of-the-Art.} In \Tref{tbl:lvu}, we comprehensively compare our approach with state-of-the-art methods~\cite{feichtenhofer2019slowfast, sun2019videobert, radford2021learning, wu2021towards, islam2022long, chen2022movies2scenes, sun2022long} on the LVU benchmark. As can be noticed from the table, short-term models such as VideoBERT~\cite{sun2019videobert} and CLIP~\cite{radford2021learning} generally struggle to perform well in contrast to the long-term models such as ViS4mer~\cite{islam2022long}, Movie2Scenes~\cite{chen2022movies2scenes} or Ours. ViS4mer~\cite{islam2022long} proposes a transformer-based model with a state-space decoder for long video classification and trains a unique model from scratch for each task in LVU. In comparison, we only train a single linear layer, on top of the representations encoded from our pretrained model, for each task. As shown in \Tref{tbl:lvu}, our model significantly outperforms ViS4mer~\cite{islam2022long} across different tasks despite having much fewer trainable parameters. These results highlight the strong merit of the proposed pretraining strategy which enabled a robustly transferable model for various long-form video understanding tasks.

Movie2Scenes~\cite{chen2022movies2scenes} is pretrained using 30,340 movies and their associated metadata from Amazon Prime Video's \textit{internal} database. \cite{chen2022movies2scenes} adopts a transformer network to learn video representation using  movie similarity as a supervision signal.
The comparison in \Tref{tbl:lvu} shows that our method gives a competitive if not better performance despite being pretrained on a dataset that is approximately 50 times smaller than the dataset used in Movie2Scenes. This is most likely attributed to the inter-modal and cross-modal constraints which enforce our model to learn the underlying multimodal cues between different modalities \textit{within} a movie, thereby compensating for what it lacks in size. For instance, the genre of a movie can be inferred from audio signals, \eg~\textit{explosion} sounds are usually associated with action-themed movies, while \textit{unsettling} background music is usually used in horror movies. This explains the strong performance of our model in metadata prediction tasks in \Tref{tbl:lvu} even if it was pretrained without metadata information. In content understanding tasks, our method outperforms Movie2Scenes by 3\% on average. 

\begin{table}[!t]
\begin{center}
\setlength{\tabcolsep}{4.5pt}
\renewcommand{\arraystretch}{1}
\caption{\textbf{Temporal event localization} on MUSES benchmark.}
\vspace{1mm}
\mytabular{
\begin{tabular}{lcccccc}
\toprule
Method & $\mathrm{mAP}_{0.3}$ & $\mathrm{mAP}_{0.4}$ & $\mathrm{mAP}_{0.5}$ & $\mathrm{mAP}_{0.6}$ & $\mathrm{mAP}_{0.7}$ & mAP   \\ \midrule
Random & 1.20 & 0.64 & 0.29 & 0.10 & 0.03 & 0.45\\
P-GCN~\cite{zeng2019graph}  & 19.9 & 17.1 & 13.1 & 9.7 & 5.4 & 13.0\\
Liu \etal ~\cite{liu2021multi}  & 26.5 & 23.1 & 19.7 & 14.8 & 9.5 & 18.7\\ \midrule
Ours & \textbf{29.5} & \textbf{26.9} & \textbf{22.8} & \textbf{16.6} & \textbf{13.3} & \textbf{21.8}\\

\bottomrule
\end{tabular}
}

\label{tbl:muses}
\end{center}
\vspace{-4mm}
\end{table}

\subsection{Transferable Model for Movie Understanding}
\label{sec:exp_part3}
In this section, we study the capabilities of our pretrained model beyond long-form video classification.  For this purpose, we make use of five movie benchmarks related to event localization~\cite{liu2021multi}, editing~\cite{argaw2022anatomy}, and retrieval~\cite{suris2022s, bain2020condensed}. For each benchmark task, we follow the official data split and training (fine-tuning) settings.

\begin{table*}[!t]
\begin{center}
\setlength{\tabcolsep}{3pt}
\renewcommand{\arraystretch}{1}
\caption{\textbf{Cinematographic scene understanding} on AVE.}
\vspace{1mm}
\mytabular{
\begin{tabular}{lccccccccc}
\toprule
Method  & Shot Size & Shot Angle & Shot Type & Shot Motion & Shot Location & Shot Subject & Num People & Sound Source & Average \\ \midrule
CLIP~\cite{radford2021learning} & 51.3 & 54.9 & 58.0 & 37.6 & 81.0 & 42.9 & 57.2 & 43.4 & 53.3 \\
ResNet101~\cite{he2016deep} & 66.8 & 55.9 & 64.7 & 33.5 & 82.1 & 46.8 & 60.2 & 32.0 & 55.2\\
AVE~\cite{argaw2022anatomy} & 65.0 & 49.5 & \textbf{65.3} & 43.2 & 83.7 & 46.7 & 61.4 & 38.9 & 56.7 \\ \midrule
Ours & \textbf{67.4} & \textbf{57.7} & 63.8 & \textbf{46.1} & \textbf{84.2} & \textbf{51.2} & \textbf{61.8} & \textbf{50.1} & \textbf{60.3}\\
\bottomrule
\end{tabular}
}
\label{tbl:ave}
\end{center}
\vspace{-4mm}
\end{table*}

\paragraph{Multi-shot Temporal Event Localization.} Localizing events in movies is a very challenging task due to the large intra-instance variation caused by frequent shot cuts. This includes actor/scene/camera change and heavy occlusions within a single instance. We thus explore the potential of our work for temporal event localization in multi-shot videos. For this purpose, we use the multi-shot events (MUSES) benchmark~\cite{liu2021multi}. MUSES is an action localization dataset with 25 categories and contains 3,697 videos collected from more than 1000 drama episodes. Each video in the dataset is typically five-to-nineteen minutes long.

We follow the task pipeline used by the state-of-the-art method of Liu~\etal~\cite{liu2021multi}. Given a video and a set of temporal proposals, the overall process in~\cite{liu2021multi} consists of three steps,~\ie feature extraction, temporal aggregation, and proposal evaluation. The key difference between \cite{liu2021multi} and our approach is the temporal aggregation step. While~\cite{liu2021multi} trains a convolution-based temporal aggregation module to mitigate the intra-instance variation, we instead use the features encoded from the transformers in our pretrained model. 

We use the mean average precision (mAP) metric to evaluate the performance of multi-shot event localization. In~\Tref{tbl:muses}, we compare our approach and state-of-the-art methods~\cite{zeng2019graph,liu2021multi} using different threshold values of temporal intersection over union (IoU). As can be inferred from~\Tref{tbl:muses}, using our pretrained model as an encoder for the temporal aggregation step leads to a significantly better localization performance compared to the baseline model~\cite{liu2021multi}. For example, our approach achieves an average mAP of 21.8 on MUSES which is approximately 16.6\% better than the state-of-the-art method. This is mainly because our pretrained model is capable of reasoning over a multi-shot input, and hence the localization network can focus on the proposal evaluation step unlike \cite{liu2021multi} where the network also has to learn temporal aggregation from scratch. 

\paragraph{Cinematographic Scene Understanding.} In the filmmaking process, movie scenes are created by sequentially assembling shots. Thus, it is apparent that movie scene understanding should be formulated by taking the building blocks,~\ie~shots, into account. In this regard, cinematographic scene understanding aims at predicting the attributes of the shots that compose a given scene. To evaluate the transferability of our pretraining to this task, we use the anatomy of video editing (AVE) benchmark~\cite{argaw2022anatomy}. AVE introduces eight tasks related to cinematographic attributes prediction, \ie~\textit{shot size}, \textit{shot angle}, \textit{shot type}, \textit{shot motion}, \textit{shot location}, \textit{shot subject}, \textit{number of people}, \textit{sound source}. The dataset contains a total of 196,176 shots obtained from 5,591 publicly-available movie scenes.

While AVE~\cite{argaw2022anatomy} formulates shot attributes classification problem on an individual shot basis, we perform a scene-level attributes prediction instead, where we first encode a scene (shot sequence) using our pretrained model and then train a \textit{single} classifier layer on top of the encoded sequence to simultaneously predict a specific attribute for each shot in the scene. \Tref{tbl:ave} shows the comparison of our approach and competing baselines on cinematographic scene understanding. Following~\cite{argaw2022anatomy}, we use the average per-class accuracy metric for evaluation. As can be inferred from~\Tref{tbl:ave}, the shot sequence representations encoded from our pretrained model result in better performance in different tasks. These results are achieved by only training classifier layers, while  previous methods fine-tune their whole network for specific tasks. 

\paragraph{Editing Pattern Prediction.} 
Here we evaluate the proposed pretraining scheme for two tasks related to video editing, \ie~\textit{shot sequence ordering} and \textit{next shot selection}. Shot sequence ordering (SSO) is defined as a classification problem where a network is tasked to predict the order of shots given a sequence of contagious but randomly shuffled shots. Next shot selection (NSS), on the other hand, takes a partial sequence of shots as a context and predicts the most-likely next shot from a list of candidate shots. We use the AVE benchmark\cite{argaw2022anatomy}, which formulates SSO as a 6-way classification task with 3 shots at a time, and NSS as a multiple-choice problem with a sequence of 9 shots, where the first 4 shots are used as a context and the remaining 5 make the candidate list. 

We follow the task setup used by \cite{argaw2022anatomy}. Given a sequence of shots, \cite{argaw2022anatomy} extracts audio-visual features from each shot clip and concatenates the extracted features as input to the next layer. In our case, we pass the extracted sequence of features into our pretrained contextual and cross-modal transformers before concatenating them. This simple modification notably improves network performance for both shot sequence ordering and next shot selection tasks. These results indicate that the proposed pretrained framework is capable of reasoning over long sequence of videos and can capture aspects related to movie style, enabling potential applications in automated video editing.\\ 

\begin{table}[!t]
\begin{center}
\setlength{\tabcolsep}{6pt}
\renewcommand{\arraystretch}{1}
\caption{\textbf{Editing pattern prediction} on AVE.}
\vspace{1mm}
\mytabular{
\begin{tabular}{lcclc}
\toprule
\multicolumn{2}{c}{Shot Sequence Ordering} & & \multicolumn{2}{c}{Next Shot Selection}\\ \cmidrule(lr){1-2} \cmidrule(lr){3-4}
Method & Acc. & & Method & Acc. \\ \midrule
Random & 16.6 & & Random & 20.0 \\
Argaw~\etal~\cite{argaw2022anatomy} (late fusion) & 24.4 & & Cosine similarity & 13.4\\
Argaw~\etal~\cite{argaw2022anatomy} (early fusion) & 30.7 &  & Argaw~\etal~\cite{argaw2022anatomy}  & 41.4\\ \midrule
Ours & \textbf{32.5} &  & Ours & \textbf{44.2} \\
\bottomrule
\end{tabular}
}
\label{tbl:ave_seq}
\end{center}
\vspace{-4mm}
\end{table}

\paragraph{Scene-Soundtrack Selection.} Scene-soundtrack selection aims to retrieve a soundtrack that best matches the theme of a given scene and vice versa. To evaluate the performance of our approach on this task, we use the recently proposed MovieClips-based benchmark~\cite{suris2022s}. Given a scene (soundtrack), we follow~\cite{suris2022s} and compute the feature distance between each soundtrack (scene) in a pool of $N=2000$ target candidates \textit{not seen during model pretraining}. After sorting the candidates based on the distance value, we evaluate the retrieval performance using two metrics: (i). \textit{Recall@$K$} checks if the ground-truth pair is in the $K$ closest candidates and, (ii). \textit{Median Rank} returns the median value of the positions of the ground-truth pair in the sorted list of candidates across the test set. 

\Tref{tbl:m4v} shows the comparison of our work with competing baselines. The results indicate that performing zero-shot video $\rightleftharpoons$ audio retrieval using the contextual audio and video features already gives competitive results. The inferior performance compared to the state-of-the-art method, MVPt~\cite{suris2022s}, can be attributed to the fact that our base audio encoders (VGGish~\cite{hershey2017cnn} and wav2vec 2.0~\cite{baevski2020wav2vec}) are not well-suited for extracting discriminative music features, thereby making zero-shot retrieval a challenging task. We address this limitation by fine-tuning (on \cite{suris2022s}) our pretrained model using DeepSim~\cite{lee2020metric}, which extracts disentangled music tagging embeddings, as our base audio encoder, \ie~we only fine-tune the contextual video and audio transformers using inter-modal contrastive loss in \Eref{eqn:loss_v_a}. Our fine-tuned model achieves a new state-of-the-art performance on scene-soundtrack selection. 
To examine the contribution of the pretraining step in the achieved result, we train the audio and video transformers from scratch using the fine-tuning dataset. We experimentally observed that such training achieves a performance equivalent to MVPt's~\cite{suris2022s}. This confirms the importance of the pretrained model for video $\rightleftharpoons$ audio retrieval task.

\begin{table}[!t]
\begin{center}

\setlength{\tabcolsep}{6pt}
\renewcommand{\arraystretch}{1}
\caption{\textbf{Scene-soundtrack selection}}
\vspace{1mm}
\mytabular{
\begin{tabular}{lccccc}
\toprule
& \multicolumn{2}{c}{Median Rank $\downarrow$} & \multicolumn{3}{c}{Recall $\uparrow$}\\ \cmidrule(lr){2-3} \cmidrule(lr){4-6}
Method & V $\rightarrow$ A & A $\rightarrow$ V & R@1 & R@5 & R@10 \\ \midrule
Random & 1000 & 1000 & 0.05 & 0.25  & 0.50 \\
MVPt~\cite{suris2022s} & 21 & 21 & 15.03 & 25.74 & 36.56\\ \midrule

Ours (zero-Shot) & 33 & 29 & 12.54 & 23.03 & 33.61\\
Ours (fine-tuned) & \textbf{13} & \textbf{10} & \textbf{15.72} & \textbf{34.96} & \textbf{46.80} \\
\bottomrule
\end{tabular}
}
\label{tbl:m4v}
\end{center}
\vspace{-4mm}
\end{table}

\paragraph{Scene Description Retrieval.}
Given a high-level semantic description of a scene, this task targets to retrieve the correct scene over all possible candidates in a dataset. To evaluate our model on this task, we use the Condensed movies (CDM) benchmark~\cite{bain2020condensed} which contains a total of 33,976 scenes and their corresponding textual description. At test time, a feature embedding from a given text query is compared with 6,581 scene embeddings out of which only one is the correct pair. We compare the performance of our approach and several state-of-the-art methods~\cite{liu2019end,miech2018learning,bain2020condensed} in different metrics as summarized in \Tref{tbl:condensed_movies}. For zero-shot scene description retrieval, we use the outputs of the contextual video and language transformers. 

As can be seen from \Tref{tbl:condensed_movies}, our pretrained model gives a competitive performance compared to previous approaches, even though it hasn't been specifically optimized for this task. We also experimented with fine-tuning the video and language branches of our model using the CDM~\cite{bain2020condensed} train set. Our fine-tuned model outperforms the previous state-of-the-art, CDM~\cite{bain2020condensed}, by a significant margin. The experimental results in \Tref{tbl:m4v} and \Tref{tbl:condensed_movies} highlight 
the flexibility of the proposed approach, where different components of our framework can be either deployed directly or further fine-tuned for specific tasks.

\begin{table}[!t]
\begin{center}

\setlength{\tabcolsep}{4.5pt}
\renewcommand{\arraystretch}{1}
\caption{\textbf{Scene description retrieval} on CDM benchmark}
\vspace{1mm}
\mytabular{
\begin{tabular}{lccccc}
\toprule
Method & R@1 & R@5 & R@10 & Median R. & Mean R.  \\ \midrule
Random & 0.01 & 0.08 & 0.15 & 3290 & 3290.0\\
CDM~\cite{bain2020condensed} & 5.6 & 17.6 & 26.1 & 50 & 243.9\\ \midrule

Ours (zero-shot) & 5.4 & 17.3 & 25.2 & 56 &  254.7\\
Ours (fine-tuned) & \textbf{7.7} & \textbf{23.1} & \textbf{32.8} & \textbf{27} & \textbf{176.3} \\
\bottomrule
\end{tabular}
}
\label{tbl:condensed_movies}
\end{center}
\vspace{-4mm}
\end{table}
\section{Conclusion}
We introduced a new pretraining strategy that leverages multimodal cues over long-range videos. We designed a model that incorporates contextual transformer layers for each modality (audio, video, language) and cross-modal transformers to capture long-range dependencies in movie clips. We trained the model in a modest dataset of movies to later test it on six different movie understanding benchmarks. Our extensive experimental results empirically demonstrated the effectiveness of our long-range multimodal pretraining strategy.

\paragraph{Acknowledgement.} This work was supported by the Institute of Information and communications Technology Planning and evaluation (IITP) grant funded by the Korea government (MSIT, No. 2021-0-02068, Artificial Intelligence Innovation Hub) and National Research Foundation of Korea (NRF) grant funded by the Korea government (MSIT, No. RS-2023-00212845).

{\small
\bibliographystyle{ieee_fullname}
\bibliography{egbib}
}

\end{document}